\documentclass[11pt]{article}
\usepackage{amsfonts}
\usepackage{amsmath}
\usepackage{amssymb}
\usepackage{graphics,graphicx}
\usepackage{hyperref}
\usepackage{subfigure}
\usepackage{algpseudocode}
\usepackage{algorithm}
\usepackage{mathtools}
\usepackage{hyperref}
\usepackage{scrextend}
\usepackage{float}
\usepackage[numbers]{natbib}
\usepackage[top=1in,bottom=1in,right=1in,left=1in]{geometry}
\usepackage{setspace}
\usepackage{endfloat}
\usepackage[titletoc]{appendix}
\usepackage{array}
\newcolumntype{L}[1]{>{\raggedright\let\newline\\\arraybackslash\hspace{0pt}}m{#1}}
\newcolumntype{C}[1]{>{\centering\let\newline\\\arraybackslash\hspace{0pt}}m{#1}}
\newcolumntype{R}[1]{>{\raggedleft\let\newline\\\arraybackslash\hspace{0pt}}m{#1}}

\begin{document}

\title{Boosted Sparse Non-linear Distance Metric Learning}
\author{
  Yuting Ma\\
  \texttt{yma@stat.columbia.edu}\\
  Department of Statistics\\
  Columbia University\\
  New York, NY 10027
  \and
 	Tian Zheng\\
  \texttt{tzheng@stat.columbia.edu} \\
  Department of Statistics\\
  Columbia University\\
  New York, NY 10027
}
\date{}
\maketitle
\begin{center}
\textbf{Abstract}
\end{center}
This paper proposes a boosting-based solution addressing metric learning problems for high-dimensional data. Distance measures have been used as natural measures of (dis)similarity and served as the foundation of various learning methods. The efficiency of distance-based learning methods heavily depends on the chosen distance metric. With increasing dimensionality and complexity of data, however, traditional metric learning methods suffer from poor scalability and the limitation due to linearity as the true signals are usually embedded within a low-dimensional nonlinear subspace. In this paper, we propose a nonlinear sparse metric learning algorithm via boosting. We restructure a global optimization problem into a forward stage-wise learning of weak learners based on a rank-one decomposition of the weight matrix in the Mahalanobis distance metric. A gradient boosting algorithm is devised to obtain a sparse rank-one update of the weight matrix at each step. Nonlinear features are learned by a hierarchical expansion of interactions incorporated within the boosting algorithm. Meanwhile, an early stopping rule is imposed to control the overall complexity of the learned metric. As a result, our approach guarantees three desirable properties of the final metric: positive semi-definiteness, low rank and element-wise sparsity. Numerical experiments show that our learning model compares favorably with the state-of-the-art methods in the current literature of metric learning.



\section{INTRODUCTION}
Beyond its physical interpretation, distance can be generalized to quantify the notion of similarity, which puts it at the heart of many learning methods, including the $k$-Nearest Neighbors ($k$NN) method, the $k$-means clustering method and the kernel regressions. The conventional Euclidean distance treats all dimensions equally. With the growing complexity of modern datasets, however, Euclidean distance is no longer efficient in capturing the intrinsic similarity among individuals given a large number of heterogeneous input variables. This increasing scale of data also poses a curse of dimensionality such that, with limited sample size, the unit density of data points is largely diluted, rendering high variance and high computational cost for Euclidean-distance-based learning methods. On the other hand, it is often assumed that the true informative structure  with respect to the learning task is embedded within an intrinsic low-dimensional manifold \cite{johnson1984extensions}, on which model-free distance-based methods, such as $k$NN, are capable of taking advantage of the inherent structure. It is therefore desirable to construct a generalized measure of distance in a low-dimensional nonlinear feature space for improving the performance of classical distance-based learning methods when applied to complex and high dimensional data. 

We first consider the Mahalanobis distance as a generalization of the Euclidean distance. Let $\{\mathbf{x}_1, \mathbf{x}_2, \dots, \mathbf{x}_n\}$ be a set of points in a feature space $\mathcal{X} \subseteq \mathbb{R}^p$. The Mahalanobis distance metric parameterized by a weight matrix $W$ between any two points $\mathbf{x}_i$ and $\mathbf{x}_j$  is given by:
\begin{equation} \label{eq: Mahalanobis_dist}
d_W(\mathbf{x}_i, \mathbf{x}_j) = \sqrt{(\mathbf{x}_i - \mathbf{x}_j)^T W (\mathbf{x}_i - \mathbf{x}_j)},
\end{equation}
where $W \in \mathbb{R}^{p \times p}$ is symmetric positive semi-definite (PSD), denoted as $W \succeq 0$. The Mahalanobis distance can also be interpreted as the Euclidean distance between the points linearly transformed by $L$:
\begin{equation} \label{eq: MahalanobisDist2}
d_W(\mathbf{x}_i, \mathbf{x}_j) =  ||L(\mathbf{x}_i - \mathbf{x}_j)||_2,
\end{equation}
where $LL^T = W$ can be found by the Cholesky Decomposition. From a general supervised learning perspective, a ``good'' Mahalanobis distance metric for an outcome $y$ at $\mathbf{x}$ is supposed to draw samples with similar $y$ values closer in distance based on $\mathbf{x}$, referred to as the \textit{similarity objective}, and to pull dissimilar samples further away, referred to as the \textit{dissimilarity objective}, in the projected space.

There has been considerable research on the data-driven learning of a proper weight matrix $W$ for the Mahalanobis distance metric in the field of \textit{distance metric learning}. Both accuracy and efficiency of distance-based learning methods can significantly benefit from using the Mahalanobis distance with a proper $W$ \cite{compsurvey}. A detailed comparison with related methods is presented in Section \ref{related_work}. While existing algorithms for metric learning have been shown perform well across various learning tasks, each is not sufficient in dealing with some basic requirements collectively. First, a desired metric should be flexible in adapting local variations as well as capturing nonlinearity in the data. Second, in high-dimensional settings, it is preferred to have a sparse and low-rank weight matrix $W$ for better generalization with noisy inputs and for increasing interpretability of the fitting model. Finally, the algorithm should be efficient in preserving all properties of a distance metric and be scalable with both sample size and the number of input variables. 

In this paper, we propose a novel method for a local sparse metric in a nonlinear feature subspace for binary classification, which is referred to as \textit{sDist}. Our approach constructs the weight matrix $W$ through a gradient boosting algorithm that produces a sparse and low-rank weight matrix in a stage-wise manner. Nonlinear features are adaptively constructed within the boosting algorithm using a hierarchical expansion of interactions. The main and novel contribution of our approach is that we mathematically convert a global optimization problem into a sequence of simple local optimization via boosting, while efficiently guaranteeing the symmetry and the positive semi-definiteness of $W$ without resorting to the computationally intensive semi-definite programming. Instead of directly penalizing on the sparsity of $W$, \textit{sDist} imposes a sparsity regularization at each step of the boosting algorithm that builds a rank-one decomposition of $W$. The rank of the learned weight matrix is further controlled by the sparse boosting method proposed in \cite{sparseboosting}. Hence, three important attributes of a desirable sparse distance metric are automatically guaranteed in the resulting weight matrix: positive semi-definiteness, low rank and element-wise sparisty. Moreover, our proposed algorithm is capable of learning a sparse metric on nonlinear feature space, which leads to a flexible yet highly interpretable solution. Feature selection might be carried out as a  spontaneous by-product of our algorithm that provides insights of variable importance not only marginally but also jointly in higher orders.  

Our paper is organized as follows. In Section 2 we briefly illustrate the motivation for our method using a toy example. Section \ref{boosting_metric} dissects the global optimization for linear sparse metric learning into a stage-wise learning via gradient boosting algorithm. Section \ref{section_nonlinear} extends the framework proposed in Section \ref{boosting_metric} to the nonlinear sparse metric learning by hierarchical expansion of interactions. We summarize some related works in Section \ref{related_work}. Section \ref{practical_remark} provides some practical remarks on implementing the proposed method in practice. Results from numerical experiments are presented in Section \ref{numeric_exp}. Finally, Section \ref{conclusion} concludes this paper by summarizing our main contributions and sketching several directions of future research.

\section{AN ILLUSTRATIVE EXAMPLE}
\begin{figure}[H]
\centering
\includegraphics[width=\textwidth]{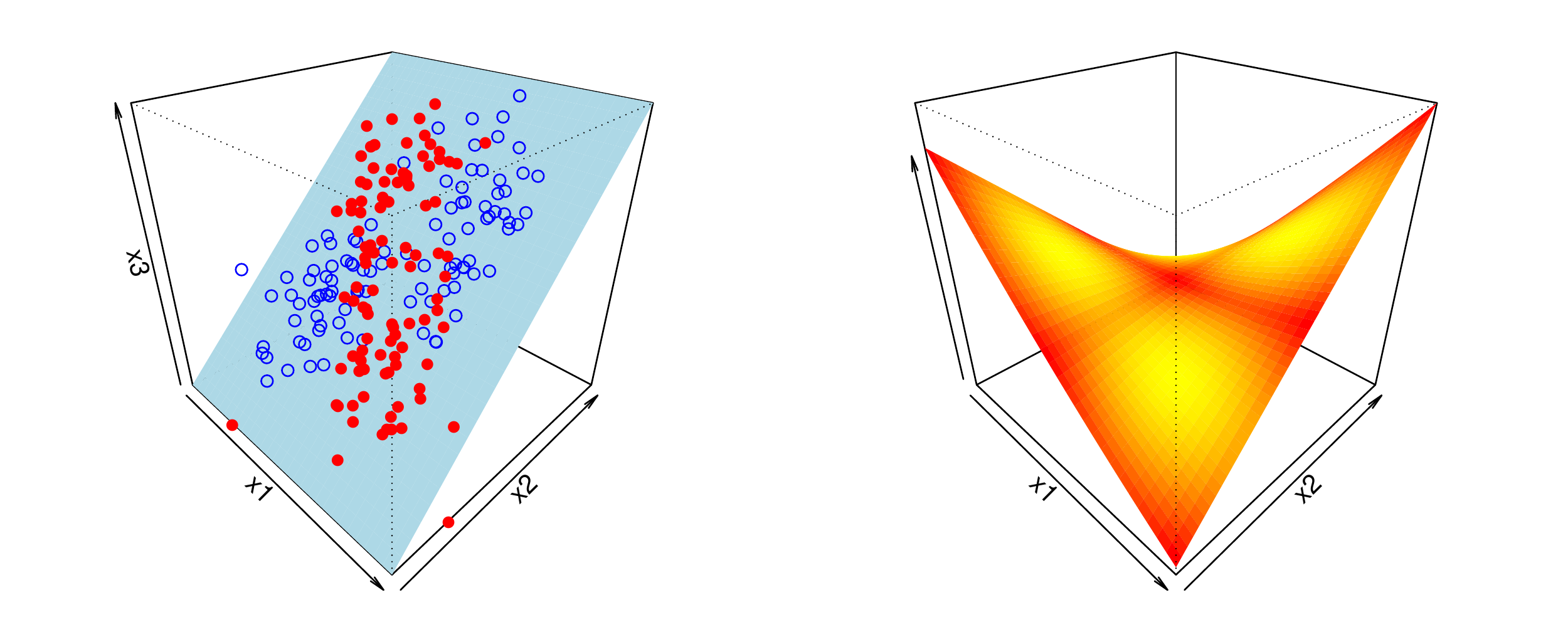}
\caption{An illustrative example of the XOR binary classification problem. \textit{Left}: Training dataset is consisted of sample points from two classes that are distributed on four clusters aligned at the crossing diagonals on a three-dimensional plane. 200 data points are generated from four bivariate Gaussian distributions and are projected into the designated  three-dimensional plane as illustrated in the figure. \textit{Right}: The  transformed subspace learned by the $sDist$ algorithm. Two horizontal dimensions are the first two input variables selected by $sDist$, that is, $x_1$ and $x_2$ in this case. The vertical dimension $z$ is the first principal component of the transformed subspace defined as $ L \phi(\mathbf{x})$, $LL^T=W$, which displays the overall shape of the surface on which new distances are computed. The colors on the grid indicates the true class probability of each class on the log scale. The yellow color indicates high probability in the class generative probability distribution and the red color indicates low probability. Since it is difficult to visualize two overlapping probability distributions in one plane, we use the same color scale for both classes and just focus on the magnitude of class generative probability distributions in each area.}
\label{fig: toy}
\end{figure}

Before introducing the details of the \textit{sDist} algorithm, we offer here a toy example in Figure \ref{fig: toy} to illustrate the problem of interest. The \textit{left} panel of Figure \ref{fig: toy} demonstrates the binary classification problem XOR (Exclusive OR) in a 3-dimensional space, which is commonly used as a classical setting for nonlinear classification in the literature. In the original space, sample points cannot be linearly separated. In this setting, sample points with the same class label are distributed in two clusters positioned diagonally from each other. In the original space, sample points cannot be linearly separated. It is also observed that the vertical dimension $x_3$ is redundant, as it provides no additional information regarding the class membership aside from $x_1$ and $x_2$. Hence, it is expected that there exists a nonlinear subspace on which points on the opposite diagonals of the tilted surface are closer to each other. Moreover, the subspace should be constructed solely based on a minimum set of variables that are informative about the class membership. The \textit{right} panel of Figure \ref{fig: toy} is the transformed subspace learned by the proposed \textit{sDist} algorithm, which is only based on the informative variables $x_1$ and $x_2$. In particular, the curved shape of the resulted surface ensures that sample points with the same class label are drawn closer and those with opposite label are pulled further apart. 

\section{BOOSTED LINEAR SPARSE METRIC LEARNING} \label{boosting_metric}
In this section, we first discuss the case of learning a linear sparse metric. Extension to nonlinear metric is discussed in Section 4. Assume that we are given a dataset $\mathcal{S} = \{\mathbf{x}_i, y_i\}$, $i=1, \dots, N$, $\mathbf{x}_i \in \mathcal{X} \subseteq \mathbb{R}^p$, where $\mathcal{X}$ is the input feature space and $p$ is the number of dimensions of the input vector\footnote{For simplicity, we only consider datasets with numerical features in this paper, on which distances are naturally defined.}. The class label $y_i \in \{-1, 1\}$. Consider an ideal scenario where there exists a metric parametrized by $W$ such that, in the $W$-transformed space, classes are separable. Then a point should, on average, be closer to the points from the same class than to the ones from the other class in its local neighborhood. Under this proposition, we propose a simple but intuitive discriminant function at $\mathbf{x}_i$ between classes characterized by $W$: 
\begin{equation} \label{eq: def_f}
f_{W, k}(\mathbf{x}_i) = d_{W,k}^-(\mathbf{x}_i) - d_{W,k}^+(\mathbf{x}_i)
\end{equation}
with 
\begin{displaymath}
\begin{aligned}
d_{W,k}^-(\mathbf{x}_i) &=  \frac{1}{k} \sum_{j \in S^-_k(\mathbf{x}_i)} (\mathbf{x}_i - \mathbf{x}_j)^T W (\mathbf{x}_i - \mathbf{x}_j) \\
d_{W,k}^+(\mathbf{x}_i) &= \frac{1}{k} \sum_{j \in S^+_k(\mathbf{x}_i)} (\mathbf{x}_i - \mathbf{x}_j)^T W (\mathbf{x}_i - \mathbf{x}_j) 
\end{aligned}
\end{displaymath}
where $S^+_k(\mathbf{x}_i)$ and $S^-_k(\mathbf{x}_i)$ are the set of $k$ nearest neighbors of $\mathbf{x}_i$ with the same class labels and with the opposite class labels as $y_i$, respectively. Without any prior information, the local neighborhoods are first identified using the Euclidean distance \footnote{In Section \ref{practical_remark}, we introduce a practical solution that updates local neighborhoods regularly as the boosting algorithm proceeds.}. When the domain knowledge of local similarity relationships are available, local neighborhoods can be constructed with better precision. The predicted class label is obtained by $\hat{y} = 1$ if $\hat{f}_W(x) >0$ and $\hat{y} = -1$ otherwise. For simplicity, we drop $k$ in the notations $f_{W, k}(\cdot)$, $d_{W,k}^-$, and $d_{W,k}^+$ as $k$ is fixed throughout the algorithm. 

The base classifier in \eqref{eq: def_f} serves as a continuous surrogate function of the $k$NN classifier, which is differentiable with respect to the weight matrix $W$. Instead of using the counts of the negative and the positive sample points in local neighborhoods, we adopt the continuous value of distances between two class to indicate the local affinity to the negative and the positive classes. Detailed comparison of the performance of the proposed classifier \eqref{eq: def_f} with the $k$NN classifier at different values of $k$ can be found in the Appendix \ref{App:AppendixA}. It is shown that $f_W$ in \eqref{eq: def_f} achieves lower test error with small values of $k$ that is commonly used in the neighborhood-based methods. Furthermore, as we will show in the following, the differentiability of $f_W$ enables smooth optimization on $W$ which facilitates a faster and more stable learning algorithm. 

Alternatively, $f_W(\mathbf{x}_i)$ can be represented as an inner product between the weight matrix $W$ and the data information matrix $D$, defined below, which contains all information of training sample point $\mathbf{x}_i$ for classification:
\begin{align} 
\hat{f}_W(\mathbf{x}_i) = \langle D_{i}, W \rangle,
\end{align}
where
\begin{equation*} 
D_{i} = \frac{1}{k} \left[ \sum_{j \in S^-_k(\mathbf{x}_i)} (\mathbf{x}_i - \mathbf{x}_j) (\mathbf{x}_i - \mathbf{x}_j) ^T - \sum_{j \in S^+_k(\mathbf{x}_i)} (\mathbf{x}_i - \mathbf{x}_j) (\mathbf{x}_i - \mathbf{x}_j) ^T \right]
\end{equation*} 
and $\langle \cdot, \cdot \rangle$ stands for the inner product for vectorized matrices. Since the matrics $D_i$'s can be pre-calcuated without the intervention of $W$, this alternative formulation of $\hat{f}_W(\mathbf{x}_i)$ suggests a computationally efficient optimization of $W$ while keeping $D_i$'s fixed. 

For learning $W$, we evaluate the performance of the classifier $f_W(\mathbf{x}_i)$ using the exponential loss, which is commonly used as a smooth objective function in binary classification:
\begin{align} \label{eq: loss_1}
L(\mathbf{y}, f_W) & = \sum\limits_{i=1}^{N}  L(y_i, f_W(\mathbf{x}_i) ) = \sum\limits_{i=1}^{N} \exp(-y_i \langle D_i, W \rangle )
\end{align}
Our learning task is then translated to derive a weight matrix $W$ on the original feature space that minimizes the loss function in \eqref{eq: loss_1}. The optimization of this objective function, however, is generally intractable for high dimensional data. Our proposed method, \textit{sDist}, seeks solution in minimizing objective function via optimizing adaptable sub-problems such that a feasible solution can be achieved. In short, the building block of \textit{sDist} are: a gradient boosting algorithm which learns a rank-one update of the weight matrix $W$ at each step; a sparsity regularization on each rank-one update to enforce the element-wise sparsity and while preserving the positive semi-definiteness simultaneously, and a sparse boosting criterion that controls the total number of boosting steps to achieve overall sparsity and low rank of the resulting weight matrix.

\subsection{Metric Learning via Boosting}
In the distance metric learning literature, much effort has been put forward to learn the weight matrix $W$ by solving a single optimization problem globally, as in \cite{DMLXing} and \cite{LMNN}. However, the optimization turns out to be either computationally intractable or susceptible to local optima with noisy high-dimensional inputs. 

Boosting \cite{freund1995boosting} offers a stagewise alternative to a single complex optimization problem. The motivation for boosting is that one can use a sequence of small improvements to derive a better global solution. Under the classification setting, boosting combines the outputs of many \textit{weak learners} trained sequentially to produce a final aggregated classifier. Here, a weak learner is a classifier that is constructed to be only modestly better than a random guess. Subsequent weak learners are trained with more weights on previously misclassified cases, which reduces dependence among the trained learners and produces a final learner that is both stable and accurate. Such an ensemble of weak learners has been proven to be more powerful than a single complex classifier and has better generalization performance \cite{ESL}. In \cite{PSDBoost} and \cite{AdaBoostDML}, a boosting algorithm has been implemented for learning a full distance metric, which has motivated the proposed algorithm in this paper. Their important theorem on trace-one semi-definite matrices is central to the theoretical basis of our approach. 



Adopting a boosting scheme, \textit{sDist} is proposed to learn a weight matrix $W$ in a stepwise fashion to avoid over-fitting to the training data in one optimization process. To construct the gradient boosting algorithm, we first decompose the learning problem into a sequence of weak learners. It is shown in \cite{PSDBoost} that for any symmetric positive semi-definite matrix $W \in \mathbb{R}^{p \times p}$ with trace one, it can be decomposed into a linear convex span of symmetric positive semi-definite rank-one matrices:
\begin{equation}
W = \sum\limits_{m=1}^{M} w_m Z_m, \quad \mbox{rank}(Z_m)=1 \mbox{ and } tr(Z_m) = 1,
\end{equation}
where $w_m \geq 0$, $m=1, \dots, M$, and $\sum\limits_{i=1}^{M} w_m= 1$. We define the vector of weights $\mathbf{w} = (w_1, w_2, \dots, w_M)$. The parameter $M \in \mathbb{Z}^+$ is the number of boosting iterations. Since any symmetric rank-one matrix can be written as an outer product of a vector to itself. We further decompose $W$ as 
\begin{equation} \label{eq: decomp_psd2}
W = \sum\limits_{m=1}^{M} w_m \xi_m \otimes  \xi_m, \quad ||\xi_m||_2 = 1 \mbox{ for all } m=1, 2, \dots, M. 
\end{equation}

Based on the decomposition in \eqref{eq: decomp_psd2}, we propose a gradient boosting algorithm that, within each step $m$, learns a rank-one matrix $Z_m = \xi_m \otimes \xi_m$ and its non-negative weight $w_m$. Each learned $Z_m$ can be considered as a small transformation of the feature space in terms of scaling and rotation. We use the following base learner in the gradient boosting algorithm:
\begin{equation} \label{eq: def_WL}
g_m(\mathbf{x}_i) = \langle D_i, Z_m \rangle.
\end{equation}

In consecutive boosting steps, the target discriminant function is constructed as a stage-wise additive expansion. At the $m^{th}$ step, the aggregated discriminant function is updated by adding the base learner $g_m(\cdot)$ with weight $w_m$ to the existing classifier with weight matrix $\hat{W}_{m-1}$ that is learned from the previous $m-1$ steps:
\begin{displaymath}
\begin{aligned}
f_{W_m}(\mathbf{x}_i) & = f_{W_{m-1}}(\mathbf{x}_i) + w_m g_m(\mathbf{x}_i)  \\
& = \langle D_i, \sum\limits_{j=1}^{m-1} w_j Z_j \rangle +  w_m \langle D_i, Z_m \rangle \\
&  =  \langle D_i, \hat{W}_{m-1} + w_m Z_m \rangle = \langle D_i, \hat{W}_m \rangle
\end{aligned}
\end{displaymath} 
where the resulting composite $\hat{W}_m$ is shown to be a weighted sum of $Z_m$'s learned from all previous steps. Therefore, the rank-one matrices obtained at each boosting step are assembled to construct the desired weight matrix, reversing the decomposition in \eqref{eq: decomp_psd2}. In this process, the required symmetry and positive semi-definiteness of weight matrix are automatically preserved without imposing any constraint. Moreover, the number of total boosting steps $M$ caps the rank of the final weight matrix. Thus, we can achieve an optimal reduced rank distance metric by using an appropriate $M$, which is discussed in Section 3.3. 

In the gradient boosting algorithm, the learning goal is to attain the minimum of the loss function in \eqref{eq: loss_1}. It is achieved by adapting a steepest-descent minimization in the functional space of $f_W$ in \eqref{eq: def_f}, which is characterized by the weight matrix $W$. The optimization problem in each boosting step is divided into two sub-steps, for $m = 1, \dots, M$: 
\begin{itemize}
\item \textbf{Finding the rank-one matrix $Z_{m}$ given the previous aggregation $\hat{W}_{m-1}$}. The residuals from the previous $m-1$ steps are:
\begin{equation} \label{eq: GBA_residual}
r_i^{(m)} = \left[- \frac{\partial L(y_i, f)}{\partial f } \right]_{f = f_{\hat{W}_{m-1}}} = y_i \exp( -y_i f_{\hat{W}_{m-1}}(\mathbf{x}_i))
\end{equation}
for $ i= 1, \dots, n$. The subsequent rank-one matrix $Z_m$ is obtained by minimizing the loss function on the current residuals for a new weak learner $g(\cdot)$ in \eqref{eq: def_WL}, that is, 
\begin{equation} \label{eq: GBA_2}
\begin{aligned}
Z_m &= \underset{Z \in \mathbb{R}^{p \times p},\  \mbox{rank}(Z) = 1}{\operatorname{arg \min}} \  \sum\limits_{i=1}^{n} L(r_i^{(m)}, g(\mathbf{x}_i))  = \underset{Z}{\operatorname{arg \min}} \sum\limits_{i=1}^{n} \exp( -r_i^{(m)} \langle D_i, Z \rangle ).
\end{aligned}
\end{equation}
Since
\begin{displaymath}
r_i^{(m)} g_m(\mathbf{x}_i) = r_i^{(m)} \langle D_i, Z_m \rangle = r_i^{(m)} \langle D_i, \xi_m \otimes \xi_m \rangle = \xi_m^T (r_i^{(m)} D_i) \xi_m, 
\end{displaymath}

the objective of \eqref{eq: GBA_2} is equivalent to identifying
\begin{equation} \label{eq: GBA_22}
\begin{aligned}
\xi_m & = \underset{\xi \in \mathbb{R}^{p},\  || \xi ||_2=1}{\operatorname{arg \min}} \  \sum\limits_{i=1}^{	n} \exp (- \xi^T  r_i^{(m)} D_i \xi),
\end{aligned}
\end{equation}
and rank-one update of weight matrix is calculated as $Z_m = \xi_m \otimes \xi_m$.


However, \eqref{eq: GBA_22} is non-convex and suffers from local minima and instability. Instead of pursuing the direct optimization on the objective function in \eqref{eq: GBA_22}, we resort to an approximation of it by the first order Taylor expansion, which is commonly used in optimizing non-convex exponential objective functions. It allow us to take advantage of the exponential loss in the binary classification task as well as avoid the expensive computational cost of considering a higher order of expansion. This approximation results in a simpler convex minimization problem : 
\begin{equation} \label{eq: GBA_pca}
\xi_m = \underset{\xi \in \mathbb{R}^{p}, \ || \xi ||_2 =1}{\operatorname{arg \min}}   -\xi^T A_m \xi 
\end{equation}
where $A_m = \sum\limits_{i=1}^{n} r_i^{(m)} D_i $. It is worthnoting that solving \eqref{eq: GBA_pca} is equivalent to computing the the eigenvector associated with the largest eigenvalue of $A_m$ via eigen-decomposition. 

\item \textbf{Finding the positive weight $w_m$ given $Z_{m}$}: 
The optimal weight in the $m^{th}$ step minimizes \eqref{eq: loss_1} given the learned $Z_m$ from the previous step. With $g_m(\mathbf{x}_i) = \langle D_i, Z_m \rangle$:
\begin{equation} \label{eq: GBA_w}
\begin{aligned}
\tilde{w}_m & = \underset{w \geq 0}{\operatorname{arg \min}} \sum\limits_{i=1}^{n} L(y_i, f_{\hat{W}_{m-1} + w Z_m}(\mathbf{x}_i)). 
\end{aligned}
\end{equation}
$\tilde{w}_m$ in \eqref{eq: GBA_w} is obtained by solving
\begin{displaymath}
\frac{\partial L}{\partial \omega} = - \sum\limits_{i=1}^{n} r_i^{(m)} g_m(\mathbf{x}_i) \exp(- w y_i g_m(\mathbf{x}_i)) = 0
\end{displaymath}
with simple algorithms such as the bisection algorithm \cite{boyd2004convex}. The vector of weights $\mathbf{w}$ is obtained by normalizing $\mathbf{w} = \frac{\tilde{\mathbf{w}}}{||\tilde{\mathbf{w}}||_2}$.
\end{itemize}
At last, the weight matrix $W_m$ is updated by 
\begin{equation} \label{eq: agg_W}
\hat{W}_m = \hat{W}_{m-1} + w_m Z_m
\end{equation}
The full algorithm is summarized in Algorithm \ref{alg: algo1} in Section 4. 

\subsection{Sparse Learning and Feature Selection}
In the current literature of sparse distance metric learning, a penalty of sparsity is usually imposed on the columns of the weight matrix $W$ or $L$, which is inefficient in achieving both element-wise sparsity and low rank in the resulting $W$. For instance, Sparse Metric Learning via Linear Programming (SMLlp) \cite{SMLlp} is able to obtain a low-rank $W$ but the resulting $W$ is dense, rendering it not applicable to high-dimensional datasets and being lack of feature interpretability. Other methods, such as Sparse Metric Learning via Smooth Optimization (SMLsm) \cite{SMLsm}, cannot preserve the positive semidefiniteness of $W$ while imposing constraints for element-wise sparsity and reduced rank. These methods often rely on the computationally intensive projection to the positive-semidefinite cone to preserve the positive semi-definiteness of $W$ in their optimization steps. With the rank-one decomposition of $W$, we achieve element-wise sparsity and low rank of the resulting weight matrix simultaneously by regularizing both $\xi$ at each boosting step and the total number of boosting steps $M$. 

First, we enforce the element-wise sparsity by penalizing on the $l_1$ norm of $\xi$. This measure not only renders a sparse linear transformation of the input space but also select a small subset of features relevant to the class difference as output at each step. The optimization in \eqref{eq: GBA_pca} is replaced by a penalized minimization problem:
\begin{equation} \label{eq: Sparse_1}
\xi_m = \underset{\xi \in \mathbb{R}^{p}, \ || \xi ||_2 =1}{\operatorname{arg \min}} \   -\xi^T A_m \xi + \lambda_{\xi} \sum\limits_{j=1}^{p} |\xi_j|
\end{equation}
where $\lambda_{\xi} > 0$ is the regularizing parameter on $\xi$.

As pointed out in Section 3.1, \eqref{eq: GBA_pca} can be solved as a eigen-decomposition problem. The optimization problem in \eqref{eq: Sparse_1}, appended with a single sparsity constraint on the eigenvector associated with the largest eigenvalue, is shown in \cite{tpower} as a sparse eigenvalue problem. We adopt a simple yet effective solution of the truncated iterative power method introduced in \cite{tpower} for obtaining the largest sparse eigenvectors with at most $\kappa$ nonzero entries. Power methods provide a scalable solution for obtaining the largest eigenvalue and the corresponding eigenvector of high-dimensional matrices without using the computationally intensive matrix decomposition. The truncated power iteration applies the hard-thresholding shrinkage method on the largest eigenvector of $A_m$, which is summarized in Algorithm \ref{alg: power_method} in Appendix \ref{App:AppendixB}.

Using parameter $\kappa$ in the sparse eigenvalue problem spares the effort of tuning the regularizing parameter $\lambda_{\xi}$ indefinitely to achieve the desirable level of sparsity. Under the context of \textit{sDist}, $\kappa$ indeed controls the level of \textit{module effect} among input variables, namely, the joint effect of selected variables on the class membership. Inputs that are marginally insignificant can have substantial influence when joined with others. The very nature of the truncated iterative power method enables us to identify informative variables in groups within each step. These variables are very likely to constitute influential interaction terms that explain the underlying structure of decision boundary which are hard to discern marginally. This characteristic is deliberately utilized in the  construction of nonlinear feature mapping adaptively, which is discussed in detail in Section \ref{section_nonlinear}. In practice, the value of $\kappa$ can be chosen based on domain knowledge, depending on the order of potential interactions among variables in the application. Otherwise, we use cross-validation to select the ratio between $\kappa$ and the number of features $p$, denoted as $\rho$, at each boosting step as it is often assumed that the number of significant features is relatively proportional to the total number of features in real applications. 

\subsection{Sparse Boosting}
The number of boosting steps $M$, or equivalently the number of rank-one matrices, bounds the overall sparsity and the rank of resulted weight matrix. Without controlling over $M$ from infinitely large, the resulted metric may fail to capture the low-dimensional informative representation of the input variable space. Fitting with infinitely many weak learners without regularization will produce an over-complicated model that causes over-fitting and poor generalization performance. Hence, in addition to sparsity control over $\xi$, we incorporate an automatic selection of the number of weak learners $M$ into the boosting algorithm by formulating it as an optimization problem. This optimization imposes a further regularization on the weight matrix $W$ to enforce a low-rank structure. Therefore, the resulting $W$ is ensured to have reduced rank if the true signal lies in a low dimensional subspace as well as guaranteeing the overall element-wise sparsity. 

To introduce the sparse boosting for choosing an $M$, we first rewrite the aggregated discriminant function at the $m^{th}$ step as a hat operator $\Upsilon_m$, mapping the original feature space to the reduced and transformed space, i.e., $\Upsilon_m: X \rightarrow \tilde{X}_m$, in which $\tilde{X}_m$ is the transformed space by $\hat{L}_m$, $\hat{L}_m \hat{L}_m^T = \hat{W}_m$. Therefore, we have
\begin{displaymath}
f_{\hat{W}_m} (x) = \langle D, \hat{W}_m \rangle = f (\Upsilon_m(X)).
\end{displaymath}
Here $\Upsilon_m$ is uniquely defined by the positive semi-definiteness of $\hat{W}_m$. Hence, we define the complexity measure of the boosting process at the $m^{th}$ step by the generalized definition of degrees of freedom in \cite{green1994nonparametric}:
\begin{equation} \label{eq: def_C}
C_m =  tr(\Upsilon_m)= tr(\hat{L}_m).
\end{equation}
With the complexity measure in \eqref{eq: def_C}, we adopt the sparse boosting strategy introduced in \cite{sparseboosting}. First, let the process carry on for a large number, $M$, of iterations; then the optimal stopping time $\hat{m}$ is the minimizer of the stopping criterion
\begin{equation} \label{eq: opt_m}
\hat{m} = \underset{1 \leq m \leq M}{\operatorname{arg \min}} \ \left\{ \sum\limits_{i=1}^{N} L(y_i,  f_{\hat{W}_m}(\mathbf{x}_i)) \right\}+  \lambda_C C_m
\end{equation}
where $\lambda_C > 0$ is the regularizing parameter for the overall complexity of $W$.

This objective is rather intuitive: $\xi_m$'s are learned as sparse vectors and thus $Z_m = \xi_m \otimes \xi_m$ has nonzero entries mostly on the diagonal at variables selected in $\xi_m$. Therefore, $tr(\hat{L_m})$ is a good approximation of the number of selected variables, which explicitly indicates the level of complexity of the transformed space at step $m$.

\section{BOOSTED NONLINEAR SPARSE METRIC LEARNING} \label{section_nonlinear}
The classifier defined in \eqref{eq: def_f} works well only when the signal of class membership is inherited within a linear transformation of the original feature space, which is rarely the case in practice. In this section, we introduce nonlinearity in metric learning by learning a weight matrix $W$ on a nonlinear feature mapping of the input variable space $\phi(\mathbf{x}): \mathcal{R}^p \rightarrow  \mathcal{R}^{\tilde{p}}$, where $\tilde{p} \geq p$. The nonlinear discriminant function is defined as
\begin{equation} \label{eq: def_f_exp}
f_W^{\phi} (\mathbf{x}_i) = \langle D_i^{\phi}, W_m \rangle
\end{equation}
where
\begin{eqnarray} \label{eq: D_phi}
D_i^{\phi} & = \frac{1}{k} \sum_{j \in S_k^-(\mathbf{x}_i)} [\phi^{(m)}(\mathbf{x}_i) - \phi^{(m)}(\mathbf{x}_j)][\phi^{(m)}(\mathbf{x}_i) - \phi^{(m)}(\mathbf{x}_j)]^T\\
& - \frac{1}{k} \sum_{j \in S_k^+(\mathbf{x}_i)} [\phi^{(m)}(\mathbf{x}_i) - \phi^{(m)}(\mathbf{x}_j)][\phi^{(m)}(\mathbf{x}_i) - \phi^{(m)}(\mathbf{x}_j)]^T \nonumber
\end{eqnarray}
Learning a ``good'' feature mapping in the infinite-dimensional nonlinear feature space is infeasible. In \cite{LMCA}, Torresani and Lee resort to the ``kernel'' trick and construct the Mahalanobis distance metric on the basis expansion of kernel functions in Reproducing Kernel Hilbert Space. Taking a different route, Kedem \textit{et al} \cite{NonlinearDML} abort the reliance on the Mahalanobis distance metric and learn a distance metric on the non-linear basis functions constructed by regression trees. Although these methods provide easy-to-use ``black box'' algorithms that offers extensive flexibility in modeling a nonlinear manifold, they are sensitive to the choices of model parameters and are subject to the risk of overfitting. The superfluous set of basis functions also hinders the interpretability of the resulting metric model with respect to the relevant factors of class separation.

In this paper, we restrict the feature mapping $\phi(\mathbf{x})$ to the space of polynomial functions of the original input variables $x_1, \dots, x_p$. The construction of nonlinear features is tightly incorporated within the boosted metric learning algorithm introduced in Section \ref{boosting_metric}. Accordingly, a proper metric is learned in concert with the building of essential nonlinear mappings suggested in the data. 

We initialized $\phi(\mathbf{x}) = (x_1, x_2, \dots, x_p)^T$ as the identity mapping at step $0$. In the following steps, based on the optimal sparse vector $\xi$ learned from the regularized optimization problem \eqref{eq: Sparse_1}, we expand the feature space by only including interaction terms and polynomial terms among the nonzero entries of $\xi$, that is, the selected features. Such strategy allows the boosting algorithm to benefit from the flexibility introduced by the polynomials without running into overwhelming computational burden and storage need. In comparison, the full polynomial expansion results in formidable increase in dimensionality of the information matrices $D^{\phi}_i$'s to as much as $(2^p)^2$.

The polynomial feature mapping also permits selection of significant nonlinear features. Kernel methods are often preferred in nonlinear classification problems due to its flexible infinite-dimensional basis functions. However, for the purpose of achieving sparse weight matrix, each basis function need to be evaluated for making the selection toward a sparse solution. Hence, using kernel methods in such a case is computationally infeasible due to its infinite dimensionality of basis functions. By adaptively expanding polynomial features, optimizing \eqref{eq: Sparse_1} on the expanded feature space is able to identify not only significant input variables but also informative interaction terms and polynomial terms. 

Before we layout the details of the adaptive feature expansion algorithm, we define the following notions: Let $\mathcal{C}_m = \{\tilde{x}_1, \dots, \tilde{x}_{p_m}\}$ be the set of candidate variables at step $m$, where $\tilde{x}$ represents the candidate feature, and $\tilde{p}_m$ is the cardinality of the set $\mathcal{C}_m$, that is, the number of features at step $m$. The set $\mathcal{C}_m$ includes the entire set of original variables as well as the appended interaction terms. Denote $\mathcal{S}_m$ as the cumulative set of the unique variables selected up to step $m$, and $\mathcal{A}_m$ be the set of variables being newly selected in step $m$. Then,
\begin{itemize}
\item[Step 0]: Set $\mathcal{C}_0 = \{\tilde{x}_1=x_1, \dots, \tilde{x}_p=x_p \}$, the set of the original variables.
\item[Step 1]: Select $\mathcal{A}_1 \subset \mathcal{C}_0$ by the regularized optimization in \eqref{eq: Sparse_1} with prespecifed $| \mathcal{A}_1 | =\kappa$. \begin{displaymath}
\mbox{Set} \quad \mathcal{S}_1 = \mathcal{A}_1; \quad \mathcal{C}_1 = \mathcal{C}_0 \cup ( \mathcal{S}_1 \otimes \mathcal{A}_1 )
\end{displaymath}
where the operator ``$\otimes$" is defined as 
\begin{displaymath}
\mathcal{S}_1 \otimes \mathcal{A}_1 = \{\tilde{x}_i \tilde{x}_j: \tilde{x}_i \in \mathcal{S}_1, \tilde{x}_j \in \mathcal{A}_1 \}
\end{displaymath}
\item[Step $m$], $m = 2, \dots, M$: Select $\mathcal{A}_m \subset \mathcal{C}_{m-1}$. Then 
\begin{equation} \label{eq: def_CC}
\mathcal{S}_m = \mathcal{S}_{m-1} \cup \mathcal{A}_m, \quad  \mathcal{C}_m = \mathcal{C}_{m-1} \cup (\mathcal{S}_m \otimes \mathcal{A}_m )
\end{equation}
\end{itemize}
Then $\phi(\mathbf{x})$ at the $m^{th}$ step of the algorithm is defined as $\phi^{(m)}(\mathbf{x}) \triangleq X_{\mathcal{C}_{m-1}}$, the vector\footnote{Here $X = [\mathbf{x}_1, \mathbf{x}_2, \dots, \mathbf{x}_n]^T$. When $\mathcal{C}$ is a set of variable or interactions of variables, $X_{\mathcal{C}}$ represents the columns of $X$ (or products of columns of $X$) listed in $\mathcal{C}$.} whose components are elements in $\mathcal{C}_{m-1}$

It is worthnoting that, in updating $D_i^{\phi^{(m)}}$, there is no need to compute the entire matrix, the cost of which is on the order of $np_{m}^3$. Instead, taking advantage of the existing $D_i^{\phi^{(m-1)}}$, it is only required to add $\delta_m \triangleq (p_m - p_{m-1})$ rows of pairwise products between the newly added terms and currently selected ones and to make the resulting matrix symmetric. The extra computational cost is reduced to $O(n \delta_m^3)$ and $\delta_m \ll p_m $ when $p$ is large. Therefore, the method of expanding the feature space in the step-wise manner is tractable even with large $p$. Since we only increase the dimension of feature space by a degree less than $\frac{1}{2}(\delta_m\kappa + \kappa)$ at each step with $M$ controlled by the sparse boosting, the proposed hierarchical expansion is computationally feasible even with high-dimensional input data. 

We integrate the adaptive feature expansion for nonlinear metric learning into the boosted sparse metric learning algorithm in Section \ref{boosting_metric}. The final algorithm is summarized in Algorithm \ref{alg: algo1}. The details of how to choose the value of parameters $\kappa$, $\lambda_C$ and $M$ are elaborated in Section \ref{practical_remark}
\begin{algorithm}[h]
\caption{\textit{sDist}: Boosted Nonlinear Sparse Metric Learning}
\begin{algorithmic}
\State Input Parameters: $\kappa$, $M$, and $\lambda_C$
\State 1) Initialization:  $\hat{W}_0 = I_{p \times p}$; $\mathcal{C}_0 = \{\tilde{x}_1=x_1, \dots, \tilde{x}_p=x_p \}$; residuals $r_i^{(0)} = y_i$, $i=1, 2, \dots, n$.
\State 2) For $m=1$ to $M$:
\State \indent (a) Define the nonlinear feature mapping $\phi^{(m)}(\mathbf{x}) = X_{\mathcal{C}_{m-1}}$; Update $D_i^{\phi^{(m)}}$ according to \eqref{eq: D_phi}
\State \indent (b) $A_m = \sum\limits_{i=1}^{n} r_i^{(m)} D_i^{\phi^{(m)}}$.
\State \indent (c) Get $\xi_m$ from the regularized minimization problem:
\begin{align}
\xi_m = \underset{\xi \in \mathbb{R}^{p_m}, || \xi ||_2 =1}{\operatorname{arg \min}}   -\xi^T A_m \xi + \lambda_{\xi} \sum\limits_{j=1}^{p_m} |\xi_j|,
\end{align}
by the truncated iterative power method (Algorithm \ref{alg: power_method} in Appendix \ref{App:AppendixB}) with corresponding $\kappa$. 
\State \indent (d) Based on the sparse solution of $\xi_m$,update $\mathcal{A}_m, \mathcal{S}_m$ and $\mathcal{C}_m$. $g_m(\mathbf{x}_i) = \xi_m^T D_i^{\phi^{(m)}} \xi_m$ for $i=1, 2, \dots, n$.
\State \indent (e) Get $w_m$ from \eqref{eq: GBA_w}  by the bisection algorithm.
\State \indent (f) Compute residuals $r_i^{(m)}$ based on \eqref{eq: GBA_residual}:
\begin{displaymath}
r_i^{(m+1)} = r_i^{(m)} \exp(-y_i \omega_m g_m(\mathbf{x}_i)), \mbox{for } i=1, \dots, n.
\end{displaymath}
\State \indent (g) Update the weight matrix:
\begin{displaymath}
\hat{W}_m = \mathcal{I}_m^T \hat{W}_{m-1} \mathcal{I}_m + w_m \xi_m \xi_m^{T}
\end{displaymath}
where $\mathcal{I}_m= (I_{p_{m-1} \times p_{m-1}}, \mathbf{0}_{p_{m-1} \times p_{m} - p_{m-1}})$, where $I_{p \times p}$ is the $p$ by $p$ identity matrix and $\mathbf{0}_{p \times q}$ is the zero matrix of dimension $p$ by $q$.
\State 3) Determine the optimal stopping time by solving
\begin{displaymath}
\hat{m} = \underset{1 \leq m \leq M}{\operatorname{arg \min}}\sum\limits_{i=1}^{N} L(y_i,  \hat{W}_m) +  \lambda_C C_m.
\end{displaymath}
Then set the output $\hat{W} = \hat{W}_{\hat{m}}$.
\end{algorithmic}
 \label{alg: algo1}
\end{algorithm}

\section{RELATED WORK} \label{related_work}
There is an extensive literature devoted on the problem of learning a proper $W$ for the Mahalanobis distance. In this paper, we focus on the problem of supervised metric learning for classification in which class labels are given in the training sample. In the following, we categorize related methods in the literature into four groups: 1) global metric learning, 2) local metric learning, 3) sparse metric learning, and 4) nonlinear metric learning. 

Global metric learning aims to learn a $W$ that addresses the similarity and dissimilarity objectives at all sample points. Probability Global Distance Metric (PGDM) learning \cite{DMLXing} is an early representative method of this group. In PGDM, the class label ($y$) is converted into pairwise constraints on the metric values between pairs of data points in the feature ($\mathbf{x}$) space: equivalence (similarity) constraints that similar pairs (in $y$) should be close (in $\mathbf{x}$) by the learned metric; and in-equivalence (dissimilarity) constraints that dissimilar ones (in $y$) should be far away (in $\mathbf{x}$). The distance metric is then derived to minimize the sum of squared distances between data points with the equivalence constraints, while maintaining a lower bound for the ones with the in-equivalence constraints. The global optimum for this convex optimization problem is derived using Semi-Definite Programming (SDP). However, the standard SDP by the interior point method requires $O(p^4)$ storage and has a worst-case computational complexity of approximately $O(p^{6.5})$, rendering it computationally prohibitive for large $p$. Flexible Metric Nearest Neighbor (FMNN) \cite{friedman1994flexible} is another method of this group, which, instead, adapts a probability framework for learning a distance metric with global optimality. It assumes a logistic regression model in estimating the probability for pairs of observations being similar or dissimilar based on the learned metric, yet suffering poor scalability as well.

The second group of methods, local metric learning methods, learn $W$ by pursuing similarity objective within the local neighborhoods of observations and a large margin at the boundaries between different classes. For examples, see the Neighborhood Component Analysis (NCA) \cite{NCA} and the Large Margin Nearest Neighbor (LMNN) \cite{LMNN}.  NCA learns a distance metric by stochastically maximizing the probability of correct class-assignment in the space transformed by $L$. The probability is estimated locally by the  Leave-One-Out (LOO) kernel density estimation with a distance-based kernel. LMNN, on the other hand, learns $W$ deterministically by maximizing the margin at class boundary in local neighborhoods. Adapting the idea of PGDM while focusing on local structures, it penalizes on small margins in distance from the query point to its similar neighbors using a hinge loss. It has been shown in \cite{LMNN} that LMNN delivers the state-of-the-art performance among most distance metric learning algorithms. Despite its good performance, LMNN and its extensions suffers from high computational cost due to their reliance on SDP similar to PGDM. Therefore, they always require data pre-processing for dimension reduction, using \textit{ad-hoc} tools, such as the Principal Component Analysis (PCA), when applied to high-dimensional data. A survey paper \cite{compsurvey} provides a more thorough treatment on learning a linear and dense distance metric, especially from the aspect of optimization. 

When the dimension of data increases, learning a full distance metric becomes extremely computationally expensive and may easily run into overfitting with noisy inputs. It is expected that a sparse distance matrix would produce a better generalization performance than its dense counterparts and afford a much faster and efficient distance calculation. Sparse metric learning is motivated by the demand of learning appropriate distance measures in high-dimensional space and can also lead to supervised dimension reduction. In the sparse metric learning literature, sparsity regularization can be introduced in three different ways: on the rank of $W$ for learning a low-rank $W$, (e.g., \cite{LMCA}, \cite{msNCA}, \cite{SMLlp}, \cite{semiSML}), on the elements of $W$ for learning an element-wise sparse $W$ \cite{logDetSML}, and the combination of the two \cite{SMLsm}. All these current strategies suffer from various limitations and computational challenges. First, a low-rank $W$ is not necessarily sparse. Methods such as \cite{SMLlp} impose penalty on the trace norm of $W$ as the proxy of the non-convex non-differentiable rank function, which usually involves heavy computation and approximation in maintaining both the status of low rank and the positive semi-definiteness of $W$. Searching for an element-wise sparse solution as in \cite{logDetSML} places the $l_1$ penalty on the off-diagonal elements of $W$. Again, the PSD of the resulting sparse $W$ is hard to maintain in a computationally efficient way. Based on the framework of LMNN, Ying \textit{et al.} \cite{SMLsm} combine the first two strategies and penalize on the $l_{(2,1)}$ norm\footnote{The $l_{(2,1)}$ norm of $W$ is given by: $||W||_{(2,1)} = \sum\limits_{h=1}^{p} (\sum\limits_{k=1}^{p} W_{hk}^2 )^{\frac{1}{2}}$ \cite{SMLsm}} of $W$ to regularize the number of non-zero columns in $W$. Huang \textit{et al.} \cite{GSML} proposed a general framework of sparse metric learning. It adapts several well recognized sparse metric learning methods with a common form of sparsity regularization $tr(SW)$, where $S$ varies among methods serving different purposes. As a limitation of the regularization, it is hard to impose further constraint on $S$ to guarantee PSD in the learned metric.


As suggested in \eqref{eq: MahalanobisDist2}, the Mahalanobis distance metric implies a linear transformation of the original feature space. This linearity inherently limits the applicability of distance metric learning in discovering the potentially nonlinear decision boundaries. It is also common that some variables are relevant to the learning task only through interactions with others. As a result, linear metric learning is at the risk of ignoring useful information carried by the features beyond the marginal distributional differences between classes. Nonlinear metric learning identifies a Mahalanobis distance metric on a nonlinear mappings of the input variables, introducing nonlinearity via well-designed basis functions on which the distances are computed. Large Margin Component Analysis (LMCA )\cite{LMCA} maps the input variables onto a high-dimensional feature space $\mathcal{F}$ by a nonlinear map $\phi: \mathcal{X} \rightarrow \mathcal{F}$, which is restricted to the eigen-functions of a Reproducing Kernel Hilbert Space (RKHS) \cite{aronszajn1950theory}. Then the learning objective is carried out using the ``kernel trick'' without explicitly compute the inner product. LMCA involves  optimizing over a non-convex objective function and is slow in convergence. Such heavy computation limits its scalability to relatively large datasets. Kedem \textit{et al.} \cite{NonlinearDML} introduce two methods for nonlinear metric learning, both of which derived from extending LMNN. $\chi^2$-LMNN uses a nonlinear $\chi^2$-distances for learning a distance metric for histogram data. The other method, GB-LMNN, exploits the gradient boosting algorithm that learns regression trees as the nonlinear basis functions. GB-LMNN relies on the Euclidean distance in the nonlinearly expanded features space without an explicit weight matrix $W$. This limits the interpretability of its results. Current methods in nonlinear metric learning are mostly based on black-box algorithms which are prone to overfit and have limited interpretability of variables.

\section{PRACTICAL REMARKS} \label{practical_remark}
When implementing Algorithm \ref{alg: algo1} in practice, the performance of the \textit{sDist }algorithm can be further improved in terms of both accuracy and computational efficiency by a few practical techniques, including local neighborhood updates, shrinkage, bagging and feature sub-sampling. We numerically evaluate the effect of the following parameters on a synthetic dataset in Section \ref{numeric_exp}.


As stated in Section \ref{boosting_metric}, the base classifier $f_{W,k}(x_i)$ in \eqref{eq: def_f} is constructed based on local neighborhoods. Without additional domain knowledge about the local similarity structure, we search for local neighbors of each sample point using the Euclidean distance. While the actual neighbors found in the truly informative feature subspace may not be well approximated by the neighbors found in the Euclidean space of all features, the learned distance metrics in the process of the boosting algorithm can be used to construct a better approximation of the true local neighborhoods. The revised local neighborhoods are helpful in preventing the learned metric from overfitting to the neighborhoods found in the Euclidean distance and thus reducing overfitting to the training samples. In practice, we update local neighborhoods using the learned metric at a number of steps in the booting algorithm. The frequency of the local neighborhood updates is determined by the trade-off between the predictive accuracy and the computational cost for re-computing distances between pairs of sample points. The actual value of updating frequency varies in real data applications and can be tuned by cross-validation. 

In addition to the sparse boosting in which the number of boosting steps is controlled, we can further regularize the learning process by imposing a shrinkage on the rank-one update at each boosting step. The contribution of $Z_m$ is scaled by a factor $0 < \nu \leq 1$ when it is added to the current weight matrix $W_{m-1}$. That is, step 2g in Algorithm \ref{alg: algo1} is replaced by
\begin{align}
\hat{W}_m = \mathcal{I}_m^T \hat{W}_{m-1} \mathcal{I}_m + \nu w_m \xi_m \xi_m^T.
\end{align}
The parameter $\nu$ can be regarded as controlling the learning rate of the boosting procedure. Such a shrinkage helps in circumventing the case that individual rank-one updates of the weight matrix fit too closely to the training samples. It has been empirically shown that smaller values of $\nu$ favor better generalization performance and require correspondingly larger values of $M$ \cite{friedman2001greedy}. In practice, we use cross-validation to determine the value of $\nu$.

\textit{\textbf{B}ootstrap \textbf{Ag}gregat\textbf{ing}} (bagging) has been demonstrated to improve the performance of a noisy classifier by averaging over weakly correlated classifiers \cite{ESL}. Correlations between classifiers are diminished by random subsampling. In the gradient boosting algorithm of $sDist$, we use the same technique of randomly sampling a fraction $\eta$\footnote{The parameter $\eta$ is referred as the ``bagging fraction'' in the following.}, $0 < \eta \leq 1$, of the training observations to build each weak learner for learning the rank-one update. This idea has been well exploited in \cite{friedman2002stochastic} with tree classifiers, and it is shown that both accuracy and execution speed of the gradient boosting can be substantially improved by incorporating randomization into the procedure. The value of $\eta$ is usually taken to be 0.5 or smaller if the sample size is large, which is tuned by cross-validation in our numerical experiments. In particular to our algorithm, bagging substantially reduces the training set size for individual rank-one updates so that $D_i$ can be computed on the fly more quickly without being pre-calculated, avoiding the need of computational memory. As a result, in applications with large sample sizes, bagging not only benefits the test error but also improves computational efficiency. 

In high-dimensional applications, it is likely that the input variables are correlated, which translates to high variance in the estimation. As \textit{sDist} can be viewed as learning an ensemble of nonlinear classifiers, high correlation among features can deteriorate the performance of the aggregated classifier. To resolve it, we employ the same strategy as in random forests \cite{breiman2001random} of random subsampling on features to reduce the correlation among weak learners without greatly increasing the variance. At each boosting step $m$, we randomly select a subset of features of size $\tilde{p}_m$ from the candidate set $C_m$, where $\kappa < \tilde{p}_m \leq p_m$, on which $D_i$'s is constructed with dimension $\tilde{p}_m \times \tilde{p}_m$. The optimization in \eqref{eq: Sparse_1} is then executed on a much smaller scale and select $\kappa$ significant features from the random subset. As with bagging, feature subsampling enables fast computation of $D_i$'s without pre-calculation. We use $\tilde{p}_m = \sqrt{p_m}$ at the $m^{th}$ boosting step, which is suggested in \cite{breiman2001random}. Although feature subsampling will reduce the chance of selecting the significant features at each boosting step, it shall be emphasized that bagging on training samples and feature subsampling should be accompanied by shrinkage and thus more boosting steps correspondingly. It is shown in \cite{ESL} that subsampling without shrinkage leads to poor performance in test samples. With sufficient number of boosting steps, the algorithm manages to identify many informative features without including a dominant number of irrelevant ones. While the actual value of $M$ depends on the applications, in general we suggest a large value of $M$ in order to cover most of the informative features in the random subsampling. Since the computational complexity of the proposed algorithm is linearly scalable in the number of boosting steps $M$ while quadratic in the feature dimension $p$, feature subsampling is more computationally efficient even with large $M$. Hence, in high-dimensional setting, reducing the dimension of feature set to $\sqrt{p}$ makes the algorithm substantially faster. Moreover, via feature subsampling, the resulting weight matrix has much less complexity measure defined in \eqref{eq: def_C} as compared to the one without feature subsampling at each boosting step. As the sparse boosting approach optimizes over a tradeoff between prediction accuracy and the complexity of the weight matrix, the resulting $W$ would still be sufficiently sparse. Therefore, the feature subsampling with large number of boosting steps does not contradict with the goal of searching for sparse solutions.

However, there is no rule of thumb for choosing the value of $M$ in advance. Since each application has different underlying structure of its significant feature subspace as well as involving with different level of noise, the actual value of $M$ varies case by case. In general, we suggest a large number of $M$, from 500 to 2000,  that is proportional the number of features $p$. When feature subsampling is applied, $M$ should be increase in an order of $\sqrt{p}$. Since the sparse boosting process is implemented, overfitting is effectively controlled even with large $M$ and thus it is recommended to start with considerably large value of $M$. Otherwise, we use cross-validation to evaluate different choices of $M$'s.

\section{NUMERICAL EXPERIMENTS} \label{numeric_exp}
In this section, we present both simulation studies and real-data applications to evaluate the performance of the proposed $sDist$ algorithm. The algorithm is implemented with the following specifications. We use 5-fold cross-validations to determine the degree of sparsity for each rank-one update $\rho$, choosing from candidate values $\{0.05, 0.1, 0.2\}$. The same cross-validation is also applied to the tune overall complexity regularizing parameter $\lambda_C  \in \{0.001, 0.01, 0.1, 1, 10\} $. In order to control the computation cost and to ensure interpretability of the selected variables and polynomial features, we impose an upper limit on the maximum order of polynomial of the expanded features. That is, when the polynomial has an order greater than a cap value, we stop adding it to the candidate feature set. For our experiments, the cap order is set to be 4. Namely, we expect to see maximally four-way interactions. The total number of boosting steps $M$ is set to be 2000 for all simulation experiments. While by sparse boosting, the actual numbers of weak learners used vary from case to case. Throughout the numerical experiments, the reported test errors are estimated using the $k$-Nearest Neighbor classifier with $k=3$ under the tuned parameter configuration. 
 
The performance of $sDist$ is compared with several other distance metric learning methods, with the $k$-Nearest Neighbor ($k$NN) representing the baseline method with no metric learning, Probability Global Distance Metric (PGDM)\cite{DMLXing},  Large Margin Nearest Neighbor (LMNN) \cite{LMNN}, Sparse Metric Learning via Linear Programming (SMLlp) \cite{SMLlp}, and Sparse Metric Learning via Smooth Optimization (SMLsm) \cite{SMLsm}. PGDM \footnote{Source of Matlab codes:  \url{http://www.cs.cmu.edu/\%7Eepxing/papers/Old_papers/code_Metric_online.tar.gz}} \cite{DMLXing} is a global distance metric learning method that solves the optimization problem:
\begin{align*}
\underset{W \succeq 0}{\operatorname{\min}}  \quad & \sum_{y_i = y_j} (\mathbf{x}_i - \mathbf{x}_j)^T W (\mathbf{x}_i - \mathbf{x}_j) \\
\mbox{s.t.} \quad & \sum_{y_i \neq y_l} (\mathbf{x}_i - \mathbf{x}_l)^T W (\mathbf{x}_i - \mathbf{x}_l) \geq 1.
\end{align*}
LMNN \footnote{Source of Matlab codes: \url{ http://www.cse.wustl.edu/~kilian/code/code.html}} learns the weight matrix $W$ by maximizing the margin between classes in local neighborhoods with a semi-definite programming. That is, $W$ is obtained by solving:
\begin{align*}
\underset{W \succeq 0, \ \xi_{ijl} \geq 0}{\operatorname{\min}}  \quad & (1-\mu) \sum\limits_{i=1}^{n} \sum\limits_{j \in \mathcal{S}^+_k(\mathbf{x}_i)} (\mathbf{x}_i - \mathbf{x}_j)^T W (\mathbf{x}_i - \mathbf{x}_j) + \mu \sum\limits_{i=1}^{n} \sum\limits_{j \in \mathcal{S}^+_k(\mathbf{x}_i)} \sum\limits_{l \in \tilde{\mathcal{S}}^-(\mathbf{x}_i)} \xi_{ijl},\\
\mbox{s.t.} \quad & (\mathbf{x}_i - \mathbf{x}_l)^T W(\mathbf{x}_i - \mathbf{x}_l) - (\mathbf{x}_i - \mathbf{x}_j)^T W (\mathbf{x}_i - \mathbf{x}_j) \geq 1 - \xi_{ijl},
\end{align*}
 where $\xi_{ijl}$'s are slack variables and $\tilde{\mathcal{S}}^-(\mathbf{x}_i) \triangleq \{ l | y_l \neq y_i \mbox{ and } d_I(\mathbf{x}_i, \mathbf{x}_l) \leq \underset{j \in \mathcal{S}^+_k(\mathbf{x}_i)}{\operatorname{\max}} d_I(\mathbf{x}_i, \mathbf{x}_j) \}$. In the experiments, we use $\mu = 0.5$ as suggested in \cite{LMNN}.  SMLlp aims at learning a low rank weight matrix $W$ by optimizing over the linear projection $L \in \mathbb{R}^{p \times D}$ with $D \leq p$ in \eqref{eq: MahalanobisDist2}:
 \begin{align*}
 \underset{L \in \mathbb{R}^{p \times D}, \ \xi_{ijl} \geq 0}{\operatorname{\min}}  \quad & \sum\limits_{(i, j, l) \in \mathcal{T}} \xi_{ijl} + \mu \sum\limits_{r=1}^{p} \sum\limits_{s=1}^{D} |L_{rs}|,\\
 \mbox{s.t.} \quad & \| L\mathbf{x}_i - L\mathbf{x}_j \|^2_2 \leq \| L\mathbf{x}_i - L\mathbf{x}_l \|^2_2  + \xi_{ijl}, \quad \forall \ (i,j,l) \in \mathcal{T},
 \end{align*}
 where $\mathcal{T} \in \{(i,j,l) \ |  \ j = \mathcal{S}^+_1(\mathbf{x}_i), \ l = \mathcal{S}^-_1(\mathbf{x}_i) \}$. In a similar manner, SMLsm\footnote{Source of Matlab codes: \url{http://www.albany.edu/~yy298919/software.html}} learns a low-rank weight matrix $W$ by employing a $l_{(2,1)}$ norm on the weight matrix $W$ to enforce column-wise sparsity. It is cast into the minimization problem:
  \begin{align*}
 \underset{U \in \mathcal{O}^p}{\operatorname{\min}} \   \underset{W \succeq 0, \ \xi_{ijl} \geq 0}{\operatorname{\min}}  \quad & \sum\limits_{(i, j, l) \in \mathcal{T}} \xi_{ijl} + \mu \sum\limits_{r=1}^{p} \left( \sum\limits_{s=1}^{D} W_{rs}^2 \right)^{\frac{1}{2}},\\
 \mbox{s.t.} \quad & 1+ (\mathbf{x}_i - \mathbf{x}_j)^T U^T W U (\mathbf{x}_i - \mathbf{x}_j) \leq (\mathbf{x}_i - \mathbf{x}_l)^T U^T W U (\mathbf{x}_i - \mathbf{x}_l)  + \xi_{ijl}, \quad \forall \ (i,j,l) \in \mathcal{T},
 \end{align*} 
 where $\mathcal{O}^p$ is the set of $p-$dimensional orthonormal matrices.
 
The effectiveness of distance metric learning in high-dimensional datasets heavily depends on the computational complexity of the learning method. PGDM deploys a semi-definite programming in the optimization for $W$ which is in the order of $O(p^2 + p^3 + n^2p^2)$ for each gradient update. LMNN requires a computation complexity of $O(p^4)$ for optimization. SMLsm converges in $O(p^3/\epsilon)$, where $\epsilon$ is the stopping criterion for convergence. In comparison, $sDist$ runs with a computational complexity of approximately $O(M[(\kappa p + p)\kappa \log p + np^2])$ where $M$ is the number of boosting iterations and $\kappa$ is the number of nonzero entries in rank-one updates. In practice, $sDist$ can be significantly accelerated by applying the modifications in the Section 5, in which $p$ is substituted by $\tilde{p}$ and $n$ is substituted by $\eta n$. 

\begin{figure} 
\centering
\hspace{-1cm}
\includegraphics[width=\textwidth]{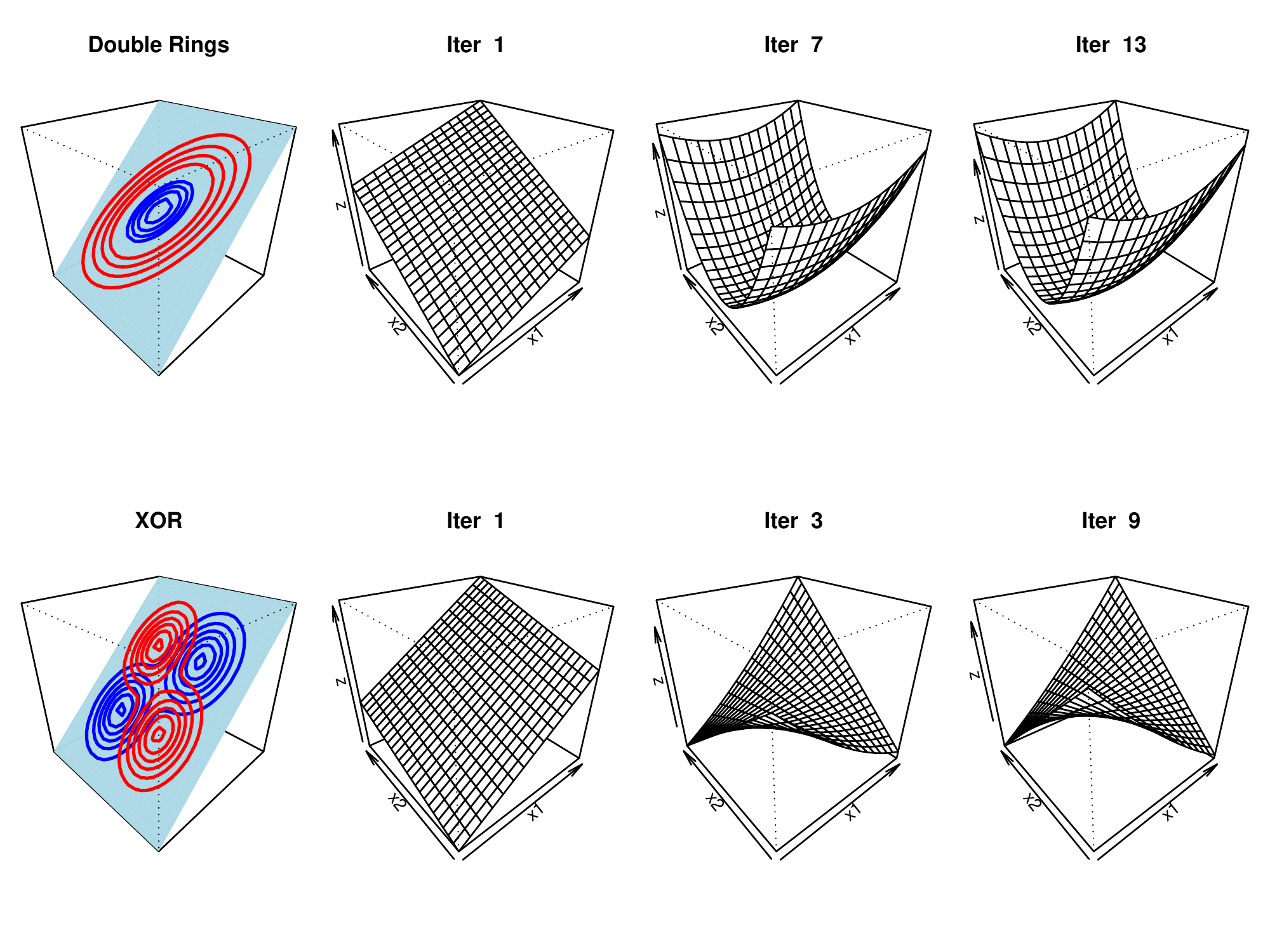}
\caption{Transformed subspaces corresponding to metrics learned for nonlinear binary classification problems. The first column shows the simulations setups. \textit{Upper}: Sample points are drawn from a ``double rings'' distribution. Shown are the contour plot of the generative class probability on a 3-dimensional surface. \textit{Lower}: Sample points are drawn from the classical XOR scenario. Columns 2-4 demonstrate how the metric learning algorithm transofrm the feature space at selected iterations. The vertical dimensions is computed as the first principle components of the transformed feature space $L \phi(\mathbf{x})$, where $LL^T=W$. Note that the solid lines in the contour plots only show the geodesic lines of high probabilities in the class generation probability distributions. The generated class labels are not separable.}
\label{fig: sim_fig}
\end{figure}

We construct two simulations settings that are commonly used as classical examples for nonlinear classification problems in the literature, the ``double ring'' case and the ``XOR'' case. In Figure 2, the left most column of the figures indicates the contour plots of high class probability for generating sample points in a 3-dimensional surface, whereas the  nput variable space is expanded to a much greater space of $p=50$, where irrelevant input variables represent pure noises. Figure \ref{fig: sim_fig} (\textit{top row}) shows a simulation study in which sample points with opposite class labels interwine in a double rings, and Figure \ref{fig: sim_fig} (\textit{bottom row}) borrows the illustrative example of ``XOR'' classification in Section 2. The columns 2-4 in Figure \ref{fig: sim_fig} illustrate the transformed subspaces learned by $sDist$ algorithm at selected iterations. Since the optimal number of iterations is not static and due to the space limit, we show only the first iteration, the last iteration determined by sparse boosting, and the middle iteration, which is rounded half of the optimal number of iterations. It is clearly shown in Figure \ref{fig: sim_fig} that the surfaces transformed by the learned distance metric correctly capture the structures of the generative distributions. In the ``double ring'' example (\textit{top row}), the learned surface sinks in the center of the plane while the rim bends upward so that sample points in the ``outer ring'' are drawn closer in the transformed surface. The particular shape owes to the quadratic polynomial of the two informative variables chosen in constructing $W$, shown as the parabola in cross-sectional grid lines. In the ``XOR'' example (\textit{bottom row}), the diagonal corners are curved toward the same directions. The interaction between the two informative variables is selected in additional to original input variable, which is essential in describing this particular crossing nonlinear decision boundary. \textit{sDist} also proves highly computationally efficient, achieving approximate optimality within a few iterations.

%

We also compare the performance of $sDist$ with other metric learning methods under different values of dimensions $p$ and sample sizes $N$ to demonstrate its scalability and its strength in obtaining essentially sparse solution in high-dimensional datasets. In this case, we generate the sample points from the ``double ring '' example and the ``XOR'' example with the numbers of informative variables being 10\% of the total dimensions, ranging from 100 to 5000. The results of these two cases are shown in Table \ref{tab: sim_DR} and Table \ref{tab: sim_XOR} respectively. It is noted that $sDist$ achieves relatively low test errors as compared to the competing methods, especially in high dimensional settings. \textit{sDist} is also proved to be scalable to datasets with large sample sizes and with high-dimensional inputs. 

\begin{table} 
\begin{tabular}{l|rrr|rrr|rrr}
\hline
\multicolumn{10}{c}{``Double Ring" Scenario }  \\
\hline
 & \multicolumn{3}{c|}{N=100} & \multicolumn{3}{c|}{N=500} & \multicolumn{3}{c}{N=5000} \\ \cline{2-10} 
\multicolumn{1}{c|}{\textbf{p}} & \multicolumn{1}{c}{50} & \multicolumn{1}{c}{500} & \multicolumn{1}{c|}{1000} & \multicolumn{1}{c}{50} & \multicolumn{1}{c}{500} & \multicolumn{1}{c|}{1000} & \multicolumn{1}{c}{50} & \multicolumn{1}{c}{500} & \multicolumn{1}{c}{1000} \\ \hline
$k$NN & 0.310 & 0.40 & 0.488 & 0.311 & 0.426 & 0.478 & 0.308 & 0.475 & 0.489 \\
PGDM & 0.320 & 0.355 & 0.389 & 0.312 & 0.356 & 0.377 & 0.337 & 0.340 & 0.412 \\
LMNN & 0.230 & 0.280 & 0.290 & 0.245 & 0.291 & 0.289 & 0.246 & 0.303 & 0.315 \\
SMLsm & 0.222 & 0.289 & 0.250 & \textbf{0.169} & 0.200 & 0.249 & 0.199 & 0.276 & 0.330 \\
sDist & \textbf{0.143} & \textbf{0.189} & \textbf{0.192} & 0.177 & \textbf{0.183} & \textbf{0.191} & \textbf{0.168} & \textbf{0.179} & \textbf{0.202} \\ \hline
Bayes Rate & 0.130 & 0.150 & 0.160 & 0.154 & 0.156 & 0.144 & 0.160 & 0.154 & 0.156 \\ \hline
\end{tabular}
\caption{Comparison of distance metric learning methods in the simulated scenario of ``Double Rings" as illustrated in Figure \ref{fig: sim_fig} (\textit{Upper} panel). Recorded are average test error over 20 simulations with varying sample size ($N$) and different total number of variables ($p$). Averaged Bayes rates are also given for reference. }
\label{tab: sim_DR}
\end{table}

\begin{table} 
\begin{tabular}{l|rrr|rrr|rrr}
\hline
\multicolumn{10}{c}{``XOR" Scenario }  \\
\hline
\textbf{} & \multicolumn{3}{c|}{N=100} & \multicolumn{3}{c|}{N=500} & \multicolumn{3}{c|}{N=5000} \\ \hline
\multicolumn{1}{c|}{\textbf{p}} & \multicolumn{1}{c}{50} & \multicolumn{1}{c}{500} & \multicolumn{1}{c|}{1000} & \multicolumn{1}{c}{50} & \multicolumn{1}{c}{500} & \multicolumn{1}{c|}{1000} & \multicolumn{1}{c}{50} & \multicolumn{1}{c}{500} & \multicolumn{1}{c}{1000} \\ \hline
$k$NN & 0.355 & 0.410 & 0.491 & 0.420 & 0.446 & 0.499 & 0.397 & 0.500 & 0.500 \\
PGDM & 0.221 & 0.355 & 0.383 & 0.289 & 0.356 & 0.360 & 0.354 & 0.350 & 0.403 \\
LMNN & \textbf{0.145} & 0.280 & 0.274 & 0.188 & 0.213 & 0.239 & 0.198 & 0.231 & 0.299 \\
SMLsm & 0.207 & 0.307 & 0.333 & 0.277 & 0.291 & 0.337 & 0.242 & 0.378 & 0.420 \\
sDist & 0.157 & \textbf{0.199} & \textbf{0.192} & \textbf{0.169} & \textbf{0.183} & \textbf{0.225} & \textbf{0.193} & \textbf{0.187} & \textbf{0.221} \\ \hline
Bayes Rate & 0.130 & 0.160 & 0.160 & 0.133 & 0.177 & 0.181 & 0.155 & 0.144 & 0.138 \\ \hline
\end{tabular}
\caption{Comparison of distance metric learning methods in the simulated scenario of ``XOR" as illustrated in Figure \ref{fig: sim_fig} (\textit{Lower} panel). Average test error is evaluated over 20 simulations with varying sample size ($N$) and different total number of variables ($p$). Averaged Bayes rates are also given for reference. }
\label{tab: sim_XOR}
\end{table}
%
%

\begin{figure} 
\centering
\hspace{-1cm}
\includegraphics[width=\textwidth]{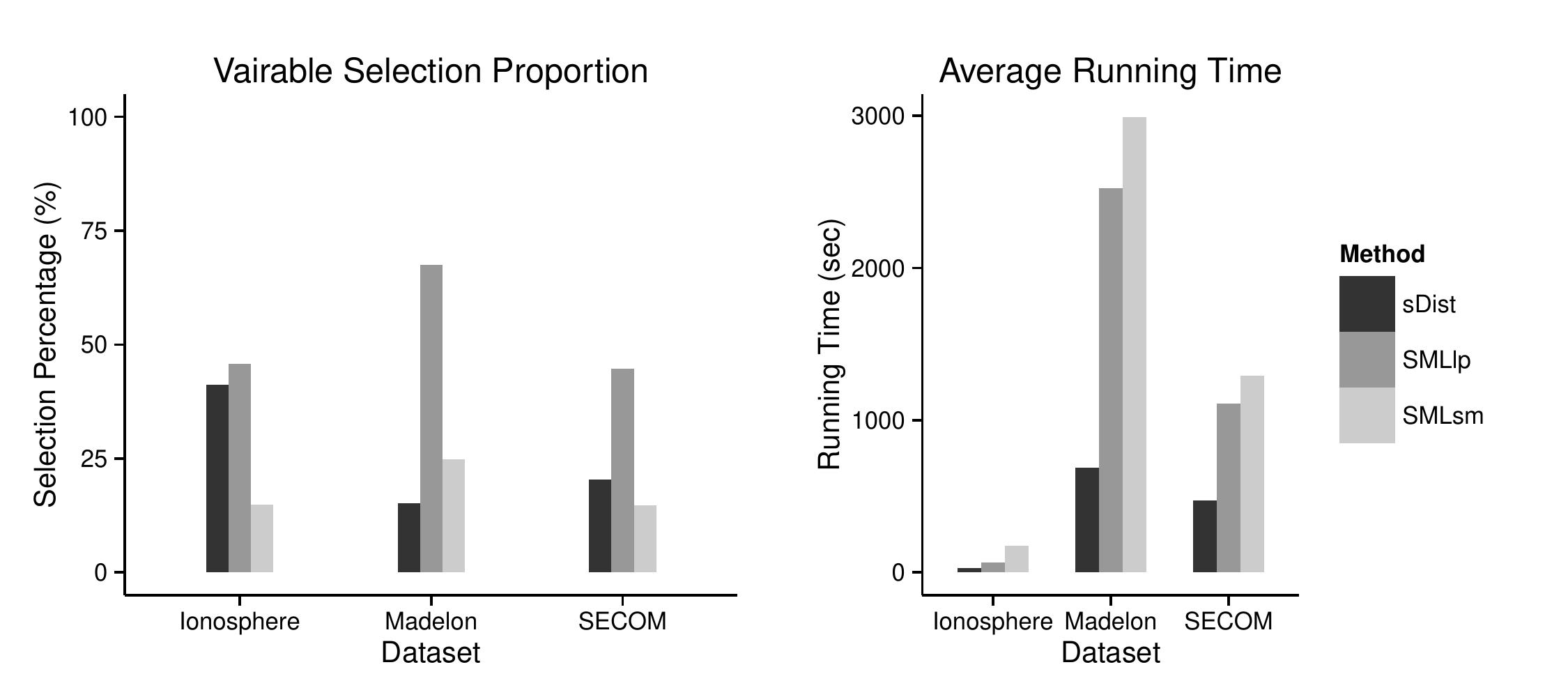}
\caption{The average percentages of variables (features) selected in the final  metrics learned by different algorithms as well as the average running times. The percentage of \textit{sDist} is calculated as the ratio of the total number of selected features over $p_{m*}$, where $p_{m*}$ is the dimension of the candidate set defined in \eqref{eq: def_C} at the optimal stopping iteration $m*$ selected by the sparse boosting method in Section 3.3.}
\label{fig: num_var}
\end{figure}

The performance of \textit{sDist} is also evaluated on three public datasets, presented in Table \ref{tab: real_data_err}. For each dataset, we randomly split the original data into a 70\% training set and a 30\% testing set, and repeat for 20 times. Parameter values are tuned by cross-validation similarly as the simulation studies. The reported test errors in Table \ref{tab: real_data_err} are the averages over 20 random splits on the datasets. The reported running times are the average CPU times for one execution\footnote{Running time of \textit{sDist} for datasets ionosphere, SECOM, Madelon are based on $M=100$, 500, and 500 respectively with the configurations that achieve the best predictive performance. The $sDist$ algorithm is implemented on R (version $3.1.3$) on x86\_64 Redhat Linux GNU system. Other competing algorithms are implemented on Matlab ($R2014a$) on the same operating system.}. We also obtain the average percentage of features selected by various sparse metric learning methods in Figure \ref{fig: num_var}. 

\begin{table}
\centering
\begin{tabular}{p{2cm} || r R{2cm} | r R{2cm}  | r R{2cm}  }
\hline
\textbf{Data Statistics }& \multicolumn{2}{c |}{Ionosphere} & \multicolumn{2}{c}{SECOM} & \multicolumn{2}{|c}{Madelon}\\
\hline
\# Inputs & \multicolumn{2}{c|}{33} & \multicolumn{2}{c}{590}  & \multicolumn{2}{|c}{500}  \\
\# Instances & \multicolumn{2}{c|}{351} & \multicolumn{2}{c}{1567} & \multicolumn{2}{|c}{2600} \\
\hline
& Test Error & Running Time (sec) & Test Error & Running Time (sec) & Test Error & Running Time (sec) \\
\hline
$k$NN & 0.13& 0.01 &0.14 &	2.07 & 0.46 &	9.05 \\
PGDM &0.07& 37.80 &0.09& 	960.47 & 0.31	& 2527.82\\
LMNN &0.06& 20.06 & 0.08 & 1960.94 & 0.39 & 1323.64 \\
SMLsm &0.09 & 173.19 & 0.09	& 1293.97 & 0.41 & 	2993.97\\
\textit{sDist} &\textbf{0.05 }&  27.49 & 	\textbf{0.07} & 473.07 & \textbf{0.09 }& 689.64 \\
\hline
\end{tabular}
\caption{Comparison of distance metric learning methods on three real public datasets. The test errors are computed using $k$-Nearest Neighbor classifier with $k=3$ based on the learned metrics from the methods under comparisons averaged over 20 random cross-validations. The recorded running times are the average CPU time for one execution.}
 \label{tab: real_data_err}
\end{table}


We first compare various distance metric learning methods on the Iononsphere dataset \cite{UCI} \footnote{Available at \tt{https://archive.ics.uci.edu/ml/datasets/Ionosphere}}. This radar dataset represents a typical small dataset. It contains mixed data types, which poses a challenge to most of the distance-based classifiers. From Table \ref{tab: real_data_err}, we see that \textit{sDist} and other metric learning methods significantly reduce the test errors by learning a nonlinear transformation of the input space, as compared to the ordinary $k$NN classifier. \textit{sDist} particularly achieves the best performance by screening out a large proportion of noises. The marginal features selected by different methods are compared in Figure \ref{fig: num_var}. Features selected by \textit{sDist} are mostly interactions within a single group of variables, suggesting an interesting underlying structure of the data for better interpretation.

SECOM \cite{UCI} \footnote{ The data is available at \url{https://archive.ics.uci.edu/ml/datasets/SECOM}. The original data is trimmed by taking out variables with constant values and variables with more than 10\% of missing values so that the dimension is reduced from 591 to 414. Observations with missing value after the trimming on variables are discarded in this experiment, which reduces the sample size to 1436.} contains measurements from sensors for monitoring the function of a modern semi-conductor manufacturing process. This dataset is a representative real-data application in which not all input variables are equally valuable. The measured signals from the sensors contain irrelevant information and high noise which mask the true information from being discovered. Under such scenario, accurate feature selection methods are proven to be effective in reducing test error significantly as well as identifying the most relevant signals \cite{UCI}. As shown in Table \ref{tab: real_data_err}, \textit{sDist} \footnote{Due to the heterogeneity in the input variables, we standardized the input variable matrix before implementing the \textit{sDist} algorithm. In the nonlinear expansions, selected interaction terms are also scaled before being added to the candidate set $\mathcal{C}$.} demonstrates dominant performance over the other three methods with an improvement about 33\% over the original $k$NN using the Euclidean distance. As compared to SMLsm, another sparse metric learning method, \textit{sDist} shows much better scalability with a large number of input variables in terms of CPU time. 

MADELON is an artificial dataset used in the NIPS 2003 challenge on feature selection \footnote{ The data is available at \url{https://archive.ics.uci.edu/ml/datasets/Madelon}. We use both the train data and the validation data. The 5-fold cross-validation is performed on the combined dataset.} \cite{UCI} \cite{guyon2006feature}\cite{guyon2007competitive}. It contains sample points with binary labels that are clustered on the vertices of a five dimensional hypercube, in which these five dimensions constitute 5 informative variables. Fifteen linear combinations of these five variables were added to form a set of 20 (redundant) informative variables while the other 480 variables have no predictive power on class label. In Table \ref{tab: real_data_err}, \textit{sDist} shows excellent performance compared to the other competing methods in terms of both predictive accuracy and computational efficiency. The test error achieved by \textit{sDist} also outperforms states-of-the-art methods beyond the distance metric learning literature on the Madelon dataset \cite{kursa2010feature} \cite{suarez2014genetic} \cite{turki2014weighted}. \textit{sDist} also attains the sparsest solution as shown in Figure \ref{fig: num_var}, with 15.2\% of features selected in the final weight matrix. Its outstanding performance indicates the importance of learning the low-dimensional manifold in high-dimensional data, particularly for the cases with low signal-to-noise ratio. 

\begin{figure}
\centering
\hspace{-1cm}
\includegraphics[width=\textwidth]{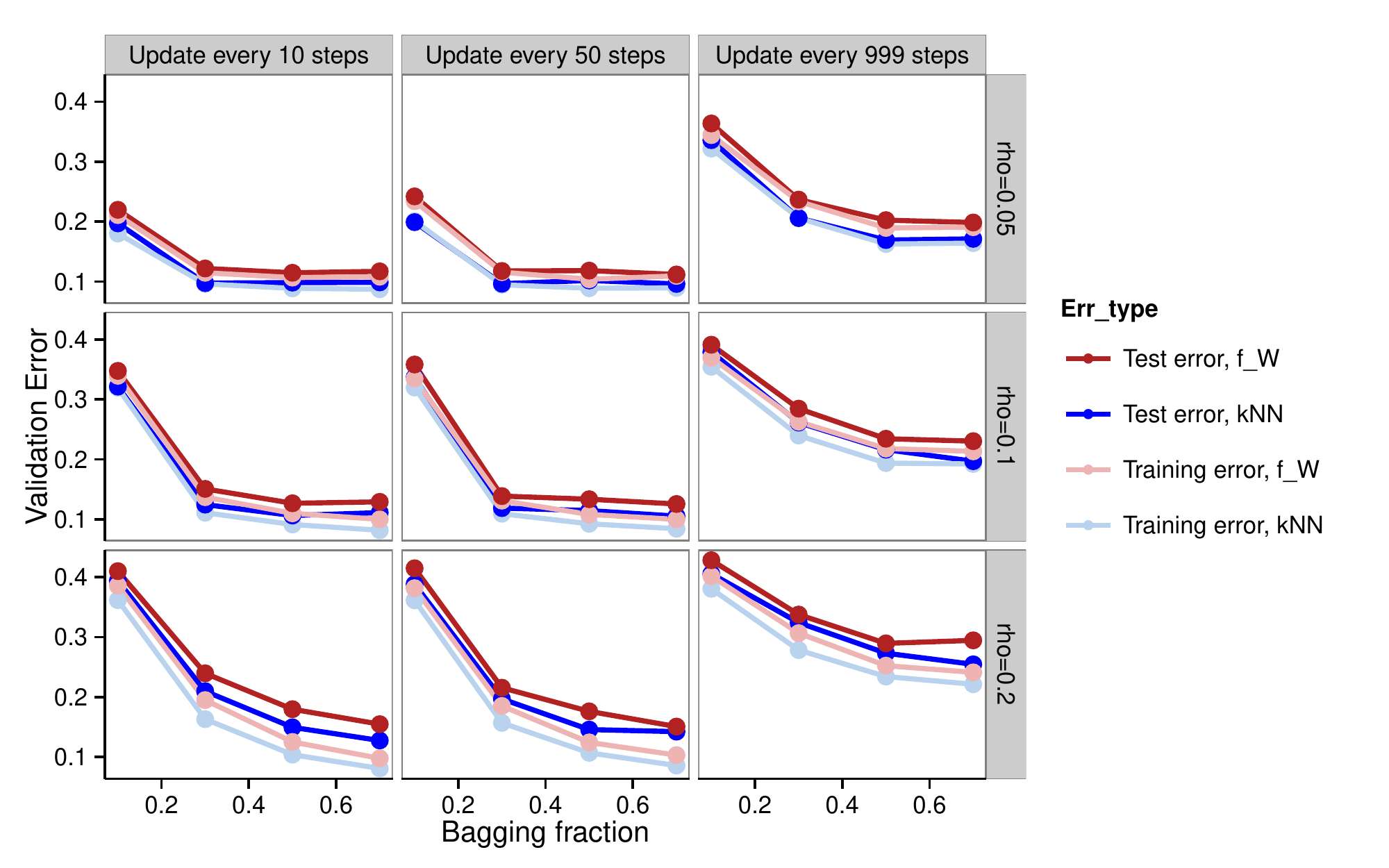}
\caption{Sensitivity analysis of  different configurations of the tuning parameters in the \textit{sDist} algorithm: frequency of local neighborhood updates, the bagging fraction $\eta$, and the degree of sparsity $\rho$ using the Madelon dataset. Training errors and testing errors are reported for both $k$NN classifier and the base classifier $f_W$ in \eqref{eq: def_f} based on 20 randomly partitioned 5-fold cross-validations.} 
 \label{fig: madelon_tune}
\end{figure}

We also experimented different configurations of the tuning parameters introduced in the algorithm and the practical remarks on the Madelon dataset, including the frequency of local neighborhood updates, bagging fraction $\eta$, and the degree of sparsity for rank-one updates $\rho$. The performances in terms of both training error and validation error are shown in Figure~\ref{fig: madelon_tune} for both the $k$NN classifier and the base classifier $f_W$ defined in \eqref{eq: def_f}. Particularly, it is evident that updating neighborhood more frequently seems to reduce validation error. The gain in performance diminishes as the frequency increases beyond a certain level. In practice, we suggest updating the local neighborhood every 50 steps as a tradeoff between the accuracy and the computational cost. In this example, the best performances of both classifiers are achieved at the bagging fraction 0.3 or 0.5 when the degree of sparsity $\rho$ is small. While as $\rho$ is large, the errors monotonically decrease as the bagging fraction increases. In practice, we suggest a bagging fraction 0.5 for moderate-size datasets and 0.3 for large datasets. When the true informative subspace is of relatively low-dimensional, as in the case of the Madelon dataset, both training errors and validation errors are reduced with small values of $\rho$.  Sparse rank-one updates benefit the most from the boosting algorithm for progressive learning and prevent overfitting at each single step,  while in other cases, the optimal value of $\rho$ depends on the underlying sparsity structure.

\section{CONCLUSIONS} \label{conclusion}
In this paper, we propose an adaptive learning algorithm for finding a sparse Mahalanobis distance metric on a nonlinear feature space via a gradient boosting algorithm for binary classification. We especially introduced sparsity control that results in automatic feature selection. The \textit{sDist} framework can be further extended in several directions. First, our framework can be generalized to multiclass problems. The base discriminant function in \eqref{eq: def_f} can be extended for a multiclass response variable in a similar fashion as in \cite{zhu2009multi} for multiclass AdaBoost. More specifically, the class label $c_i$ is recoded as a $K$-dimensional vector $\mathbf{y}_i = \{y_{i,1}, y_{i,2}, \dots, y_{i,K}\}^T$ with $K$ being the number of classes. Here $y_{ij} = 1$ if $c_i = j$ and $-\frac{1}{K-1}$ otherwise. Then a natural generalization of loss function in \eqref{eq: loss_1} is given by:
\begin{equation*}
L(\mathbf{y}, f_W^{\phi}) = \sum\limits_{i=1}^{n} \exp\left( -\frac{1}{K} \mathbf{y}_i^T f_W^{\phi}(\mathbf{x}_i) \right).
\end{equation*}
The other way is to redefine the local positive/negative neighborhood as the local similar/dissimilar neighborhood as in \citep{LMNN}, where the{ \em similar} points refer to sample points with the same class label and the {\em dissimilar} ones are those with different class labels. However, a rigorous discussion on the extension to muticlass problems requires comprehensive analysis. It is not entirely straightforward in how to exactly address multiclass labels in metric learning, or whether the learning goal is to determine a common metric for all classes or to construct different metrics between pairs of classes. Due to the limited scope of this paper, we will leave these questions in future studies. 

Furthermore, in the proposed $sDist$ algorithm, we approached the fitting of nonlinear decision boundary through interaction expansion and local neighborhoods. It has been noted that distance measures have close connections with kernel functions, which is commonly used for nonlinear learning methods in the literature. Integrating the nonlinear distance metric learning with kernel methods will lead to more flexible and powerful classifiers. 

\newpage

\begin{appendices}
\section{Comparison between the Classifier $f_W$ in \eqref{eq: def_f} and the $k$NN Classifier} \label{App:AppendixA}
\begin{figure} 
\centering
\hspace{-1cm}
\includegraphics[width=\textwidth]{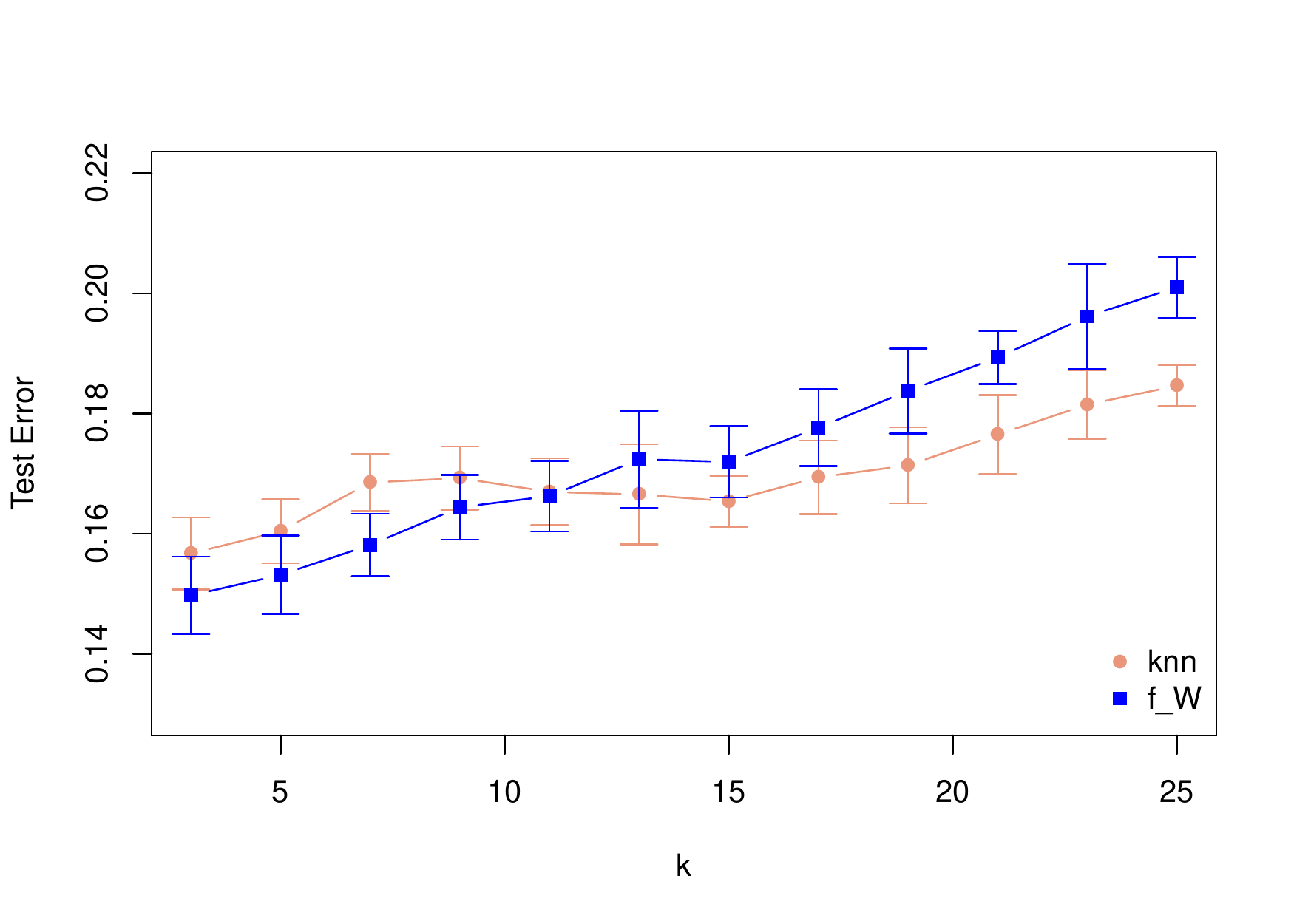}
\caption{Comparison between $f_W$ in \eqref{eq: def_f} and the $k$NN classifier in terms of the average test errors based on 20 randomly partitioned 5-fold cross-validations using the real dataset Ionosphere.} 
\label{fig: compare_knn_fw}
\end{figure}

The classifier in \eqref{eq: def_f} can be considered as a continuous surrogate function of the $k$-Nearest Neighbor classifier, which is differentiable with respect to $W$. In Figure \ref{fig: compare_knn_fw}, we show the performance of the kNN classifier and $f_W$ in \eqref{eq: def_f} at different values of $k$ on the real dataset Ionosphere. It suggests that, with small $k$ ($k \leq 11)$ which is normally used in neighborhood-based method, $f_W$ consistently outperforms $k$NN classifier with aligned pattern in terms of the average test errors based on 20 randomly partitioned cross-validations.

\section{Truncated Power Method} \label{App:AppendixB}
At each boosting step, we solve the constrained optimization problem in \eqref{eq: Sparse_1} using the truncated power method as given in Algorithm 2.
\begin{algorithm}[h] 
\caption{Truncated power method for solving \eqref{eq: Sparse_1} at the $m^{th}$ boosting step}
\begin{algorithmic}
\State Input: $A_m \in \mathbb{R}^{p_m \times p_m}$, $\kappa \in \{1, 2, \dots, p_m\}$, and the regularizing parameter $\lambda_0 > 0$
\State 1) Initialization:  $A_0 = A_m$, $\xi_0 = \frac{\mathbf{1}_{p_m}}{\sqrt{p_m}}$ 
\State 2) Iteration: For $t = 1, 2, \dots$, repeat until convergence 
\State \indent (a) Update $\lambda_t = 10\lambda_0$ until $A_t = A_{t-1} + \lambda_t I_p$ becomes positive semi-definite.
\State \indent (b) Compute $\hat{\xi}_t = \frac{A_t \xi_{t-1}}{||A_t \xi_{t-1}||}$.
\State \indent (c) Let $F_t = \mbox{supp}(\hat{\xi}_t, \kappa)$ be the indices of $\hat{\xi}_t$ with the largest $\kappa$ absolute values. Compute $\tilde{\xi}_t = \mbox{Truncate}(\hat{\xi}_t, F_t)$.
\State \indent (d)  Normalize $\xi_t = \frac{\tilde{\xi}_t}{||\tilde{\xi}_t||}$.
\State Output: $\xi_m = \xi_t$
\end{algorithmic}
\label{alg: power_method}
\end{algorithm}

It is worth noting that $A_m$ in each step of gradient boosting is not guaranteed to be positive semi-definite. Thus, to ensure that the objective function to be non-decreasing, we add a positive diagonal matrix $\tilde{\lambda} I_p$ to the matrix $A$ for $\tilde{\lambda}$ large enough such that $\tilde{A} = A + \tilde{\lambda}I_p$ is positive semi-definite and symmetric. Such change only adds a constant term to the objective function, which produces a different sequence of iterations, and there is a clear tradeoff. If $\tilde{\lambda}$ dominates $A$, the objective function becomes approximately a squared norm, and the algorithm tends to terminate in only a few iterations. In the limiting case of $\tilde{\lambda} \rightarrow \infty$, the method will not move away from the initial iterate. To handle this issue, we adapt a stochastic method that gradually increase $\tilde{\lambda}$ during the iterations and we do so only when the monotonicity is violated, as shown in the step 1 of Algorithm \ref{alg: power_method}. This truncated power method allows fast computation of the largest $\kappa$-sparse eigenvalue. For s high-dimensional but sparse matrix $A_m$, it also supports sparse matrix computation, which decreases the complexity from $O(p^3)$ to $O(\kappa p T)$, where $T$ is the number of iterations. 

\end{appendices}

\newpage
\bibliographystyle{ieeetr}
\bibliography{sDist_SADM_firstRevision.bib}

\end{document}